\documentclass[10pt,twocolumn,letterpaper]{article}

\usepackage{iccv}
\usepackage{times}
\usepackage{epsfig}
\usepackage{graphicx}
\usepackage{url}            
\usepackage{booktabs}       
\usepackage{nicefrac}       
\usepackage{microtype}      
\usepackage{amsfonts}       
\usepackage{bm}
\usepackage{custom_macros}
\usepackage{amsmath}
\usepackage[symbol]{footmisc}
\usepackage[table]{xcolor}
\usepackage{multirow}
\usepackage{amssymb}
\usepackage{eso-pic}
\usepackage{color}
\usepackage{ulem}
\usepackage[pagebackref=true,breaklinks=true,letterpaper=true,colorlinks,bookmarks=false]{hyperref}
\usepackage[capitalize]{cleveref}

\definecolor{Gray}{gray}{0.9}
\definecolor{LightCyan}{rgb}{0.88,1,1}
\definecolor{DarkBlue}{rgb}{0,0, 0.4}

\iccvfinalcopy 

\ificcvfinal\fi

\begin{document}

\title{Unify, Align and Refine:

Multi-Level Semantic Alignment for Radiology Report Generation}

\author{Yaowei Li$^{1,2}$, \quad
Bang Yang$^{1,2}$, \quad
Xuxin Cheng$^{1}$,  \quad
Zhihong Zhu$^{1}$,  \quad
Hongxiang Li$^{1}$, \quad
Yuexian Zou$^{1,2}$\thanks{Corresponding authors.} \\[5pt]
\textsuperscript{1}School of ECE, Peking University $\quad$\textsuperscript{2}Peng Cheng Laboratory \\[5pt]
{\tt\small \{ywl, chengxx, zhihongzhu, lihongxiang\}@stu.pku.edu.cn, \{yangbang, zouyx\}@pku.edu.cn}
}

\maketitle
\ificcvfinal\thispagestyle{empty}\fi

\begin{abstract}
Automatic radiology report generation has attracted enormous research interest due to its practical value in reducing the workload of radiologists. However, simultaneously establishing global correspondences between the image (e.g., Chest X-ray) and its related report and local alignments between image patches and keywords remains challenging. 
To this end, we propose an \textbf{U}nify, \textbf{A}lign and then \textbf{R}efine (\textbf{UAR}) approach to learn multi-level cross-modal alignments and introduce three novel modules: Latent Space Unifier (LSU), Cross-modal Representation Aligner (CRA) and Text-to-Image Refiner (TIR). 
Specifically, LSU unifies multimodal data into discrete tokens, making it flexible to learn common knowledge among modalities with a shared network. 
The modality-agnostic CRA learns discriminative features via a set of orthonormal basis and a dual-gate mechanism first and then globally aligns visual and textual representations under a triplet contrastive loss. 
TIR boosts token-level local alignment via calibrating text-to-image attention with a learnable mask. 
Additionally, we design a two-stage training procedure to make UAR gradually grasp cross-modal alignments at different levels, which imitates radiologists' workflow: writing sentence by sentence first and then checking word by word.
Extensive experiments and analyses on IU-Xray and MIMIC-CXR benchmark datasets demonstrate the superiority of our UAR against varied state-of-the-art~methods.
\end{abstract}

\normalem \section{Introduction}
\label{sec:intorduction}
Automatic radiology report generation, as a potential intelligent assistant to relieve radiologists from the heavy workload, has attracted a surge of research interests in recent years~\cite{jing2018automatic, li2018hybrid, jing2019show, li2019knowledge,chen2021cross, liu2021exploring, wang2021self, you2022jpg}. Mainstream methods adopt the \textit{de facto} encoder-decoder framework, where a medical image (\eg, chest X-ray) is first encoded as latent representations via convolutional neural networks (CNNs)~\cite{KarenSimonyan2015VeryDC,he2016deep}, and then further decoded into a radiology report comprised of natural language sentences by recurrent neural networks (RNNs)~\cite{hochreiter1997long,chung2014empirical} or fully-attentive networks like Transformer~\cite{vaswani2017attention}. The key problems in this task are twofold: 1) how to obtain comprehensive information associated with the input medical image and 2) how to accurately establish cross-modal alignments (CMA), \eg, matching generated words with their corresponding image regions.

\begin{figure}[t]
\begin{center}
\includegraphics[width=1\linewidth]{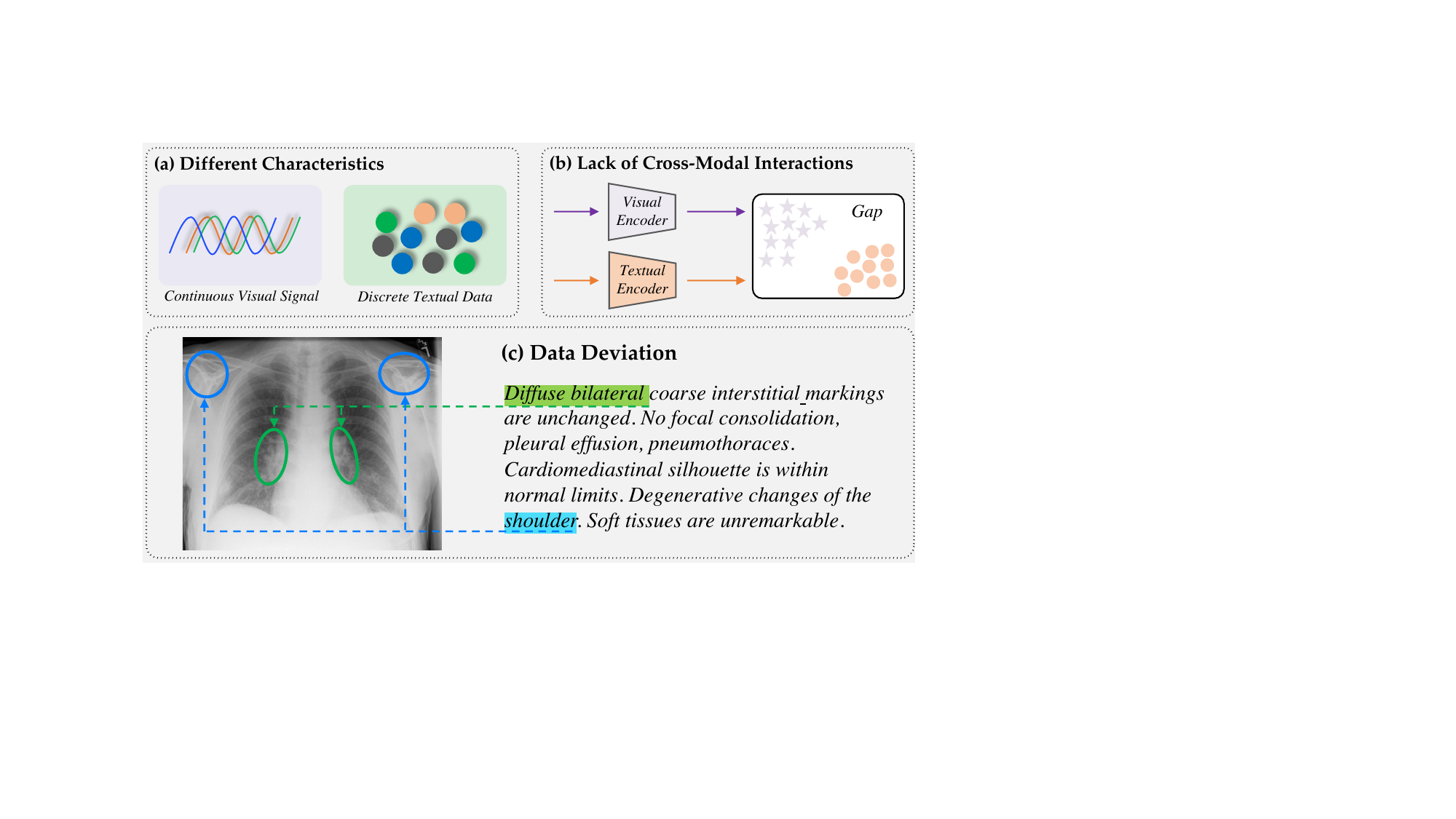}
\end{center}
\caption{
Challenges of modeling cross-modal alignments in radiology report generation. 
At the encoding stage, it is intractable to \textit{globally align} visual and textual semantics due to (a) their different characteristics (i.e., \textit{continuous} signals vs. \textit{discrete} data) and (b) the lack of cross-modal interactions. 
(c) During decoding, capturing the \textit{fine-grained alignment} between keywords and image patches is difficult because of the data deviation problem.
}

\label{fig:introduction}
\end{figure}

This paper targets at improving CMA in automatic radiology report generation, which however, is hindered by three factors shown in Figure~\ref{fig:introduction}. 
First of all, \textit{continuous} visual signals and \textit{discrete} text data have the very different characteristics, resulting in semantic inconsistency and modality-independent encoding pipelines in common practices. 
Secondly, as illustrated in Figure~\ref{fig:introduction} (b), the vision and text modalities are usually encoded via different backbones without cross-modal interactions \cite{chen2021cross, you2022jpg, qin2022reinforced}, leading to disparities in the representation space. 
Thirdly, as shown in Figure~\ref{fig:introduction} (c), there exists the so-called ``data deviation'' problem, where important details in the report (\eg, abnormalities) are sparse. As a result, the above three factors pose challenges to learning \textit{global} and \textit{local} CMA between radiographs and related reports. 
To this end, although there has emerged progresses for improving either global CMA~\cite{chen2021cross, wang2021self, you2022jpg, qin2022reinforced} or local CMA~\cite{jing2018automatic, wang2018tienet, yuan2019automatic, chen2020generating}, how to exploit multi-level CMA to enhance radiology report generation is underexplored.

Considering the above issues, we propose a \textbf{U}nify, \textbf{A}lign and \textbf{R}efine (\textbf{UAR}) framework to facilitate multi-level CMA for generating faithful and believable radiology reports. Within UAR, we design a Latent Space Unifier (LSU) to tokenize images into discrete visual tokens with a discrete variational autoencoder (dVAE)~\cite{ramesh2021zero}. 
By doing so, LSU unifies vision and text modalities into discrete tokens, making it flexible to design a shared network that seamlessly processes both modalities to learn common knowledge \cite{wang2022unifying,you2022learning}. 
Next, we introduce a modality-agnostic Cross-modal Representation Aligner (CRA) to learn global CMA. 
In implementation, CRA not only learns discriminative visual and textual features based on a set of orthonormal basis and a dual-gate mechanism, but also globally aligns both type of features under the supervision of a triplet contrastive loss~\cite{schroff2015facenet}. 
What's more, we improve local CMA by integrating Transformer~\cite{vaswani2017attention} with our proposed Text-to-Image Refiner (TIR). 
Different from the vanilla attention mechanism in Transformer, TIR additionally contains a learnable mask to re-calibrate text-to-image attention activations. TIR constrains the learnable mask with an auxiliary loss to focus word prediction on useful visual information. 
Last but not least, we design a two-stage training procedure to make the full model grasp CMA at different levels gradually. 

We conduct experiments on two widely-adopted radiology report generation benchmarks, \ie, IU-Xray~\cite{demner2016preparing} and MIMIC-CXR~\cite{johnson2019mimic}. The results demonstrate that our UAR outperforms state-of-the-art methods by a large margin, \eg, with up to 1.9\% and 15\% absolute improvements in terms of BLEU-4~\cite{papineni2002bleu} and CIDEr~\cite{vedantam2015cider} on IU-Xray, respectively. Ablation study is also carried out to understand the effect of each module of our approach.

In brief, our contributions are three-fold:

$\bullet$ To facilitate multi-level cross-modal alignments, we propose a Unify, Align and Refine framework with three modules: Latent Space Unifier (LSU), Cross-modal Representation Aligner (CRA), and Text-to-Image Refiner (TIR).

$\bullet$ LSU unifies vision and text modalities into discrete tokens, based on which CRA learns to align the global semantics of both modalities whereas TIR encourages the token-level text-to-image alignment.

$\bullet$ Extensive experiments and analyzes on IU-Xray and MIMIC-CXR datasets validate the superiority and effectiveness of our approach, which sets state-of-the-art performance and generates radiology reports accurately.

\normalem \section{Related Work}

\subsection{Visual Captioning}
As a member of AI-generated content, neural-network-based visual captioning methods~\cite{venugopalan2015translating, vinyals2015show, johnson2016densecap, yang2021non, chen2021scan2cap} are rapidly emerging. Mainstream works use CNNs to extract global~\cite{donahue2015long, mrnn2015, donahue2015long}, grid~\cite{xu2015show, lu2017knowing, rennie2017self} or region~\cite{karpathy2015deep, anderson2018bottom, lu2018neural} features, and utilize RNNs~\cite{vinyals2015show, donahue2015long} with varied attention mechanisms~\cite{song2017hierarchical, anderson2018bottom, jiang2018recurrent, lu2018neural} to capture cross-modal interactions and generate descriptions in an auto-regressive manner. In recent years, a growing number of methods leverage the potential of the fully-attentive Transformer~\cite{vaswani2017attention} in capturing long-range dependencies to boost captioning performance~\cite{cornia2020meshed, zhou2020unified, lin2022swinbert, fang2022injecting}. However, these methods are designed to generate a one-sentence description, which is brief and could lack details. 
Although several works~\cite{krause2017hierarchical, liang2017recurrent} use hierarchical models to generate long paragraphs, they may fail to capture accurate concepts and keywords, which limits application of such models in radiology report generation.

\subsection{Chest X-ray Report Generation}
To adapt the conventional visual captioning framework to radiology report generation, various improvements has been proposed. HRGR~\cite{li2018hybrid} designs a retrieval-generation method, which decides to either generate a new sentence or retrieve an existing template for different topics. CMAS~\cite{jing2019show} utilizes two writers to describe the abnormal and normal regions respectively. PPKED~\cite{liu2021exploring} incorporates the posterior-and-prior knowledge for alleviating the data deviation problem. CA~\cite{liu2021contrastive} compares the current input image with normal images to obtain discriminative abnormal features. Furthermore, for generating more coherent and consistent reports, reinforcement learning~\cite{li2018hybrid, liu2019clinically, jing2019show, qin2022reinforced} is employed to directly optimize desired metrics, graph-based methods~\cite{zhang2020radiology, li2019knowledge} are used to construct relationships between diseases, and curriculum learning~\cite{liu2021competence} as a training strategy simulates the writing pattern of ``easy first, then difficult".

Moreover, a portion of works~\cite{jing2018automatic} focus on improving cross-modal alignments (CMA) in radiology report generation. 
CoAtt~\cite{jing2018automatic} utilizes semantic information produced by an auxiliary tag prediction task to improve fine-grained CMA.
Self-boosting~\cite{wang2021self} extracts the text-correlated visual feature via coupling the image-text matching branch. 
TieNet~\cite{wang2018tienet} employ a multi-level attention for highlighting the meaningful words and regions. 
R2GenCMN~\cite{chen2021cross}, CMM+RL~\cite{qin2022reinforced} and JPG~\cite{you2022jpg} treat trainable memory networks as the intermediary of vision and text modalities to enhance global CMA. 
However, these methods achieve CMA implicitly, which could result in insufficient and inaccurate semantic matching. 
By contrast, our approach uses explicit constraints to gradually grasp multi-level CMA.

\normalem \section{Method}

\begin{figure*}[t]
\begin{center}
\includegraphics[width=0.9\textwidth]{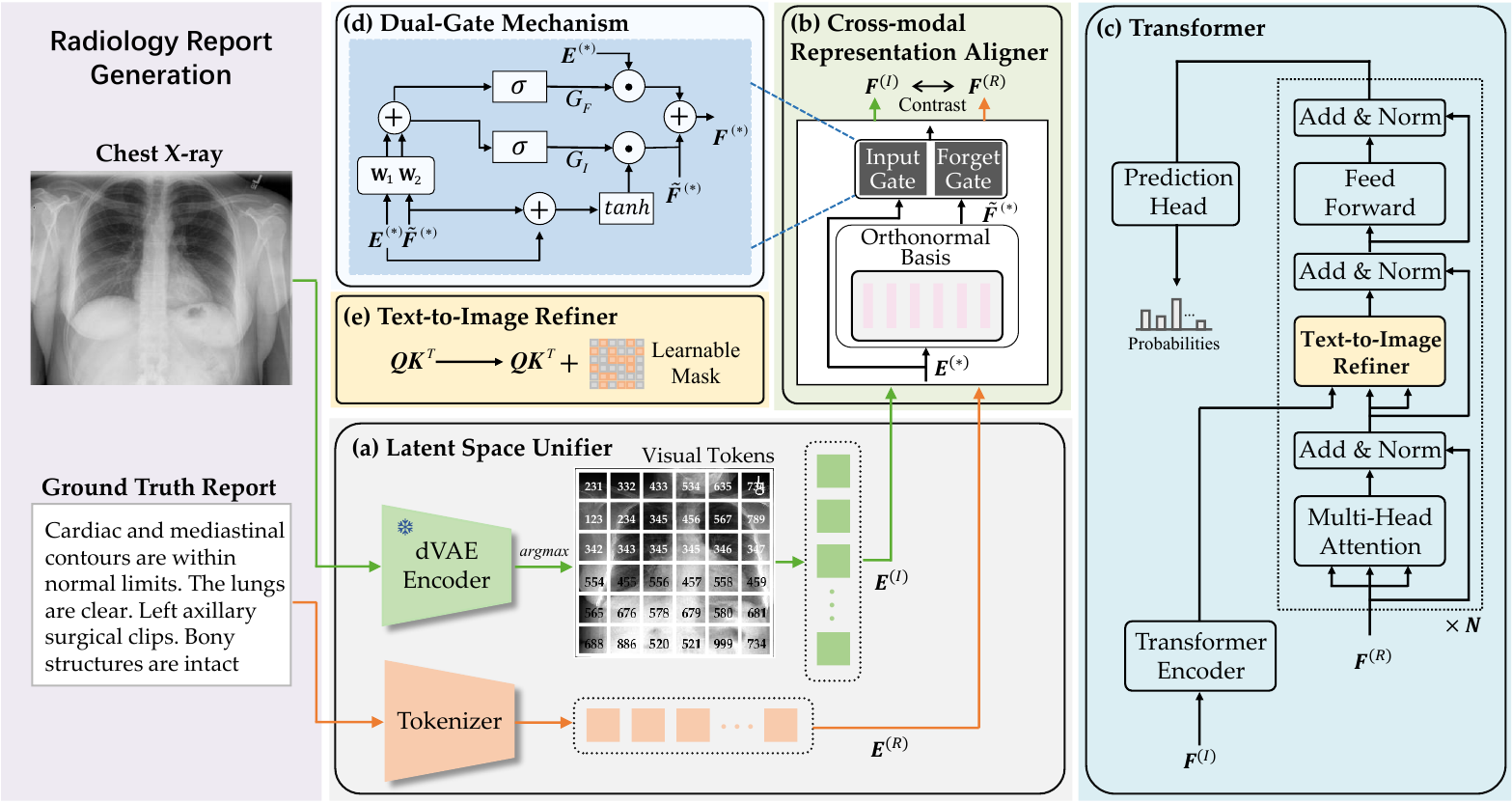} 
\end{center}
   \caption{Overview of our UAR framework. In addition to the widely-adopted Transformer shown in (c), it comprises three novel modules to boost multi-level cross-modal alignments: (a) Latent Space Unifier (\cref{sec:lsu}), (b) Cross-modal Representation Aligner (\cref{sec:cra}), and (c, e) Text-to-Image Refiner (\cref{sec:ft2tr}). In (d), we illustrate the flow chart of the dual-gate mechanism shown in (b).}

\label{fig:Framework}
\end{figure*}

\subsection{Overview}
\label{sec:overview}
As shown in Figure~\ref{fig:Framework}, our proposed UAR is built upon the encoder-decoder framework and contains three novel modules: Latent Space Unifier (LSU), Cross-modal Representation Aligner (CRA), and Text-to-Image Refiner (TIR). Given a radiograph $I$, our UAR aims to generate a descriptive report $R=\{r_1, r_2, \dots, r_T\}$ of $T$ words automatically. During training, we formulate our approach as follows:
\begin{align}
\label{eq:decompose}
&\bm{E}^{(I)}, \bm{E}^{(R)} = {\rm LSU}(I, R);\\
&\bm{F}^{(I)}, \bm{F}^{(R)} = {\rm CRA}(\bm{E}^{(I)}, \bm{E}^{(R)});\\
&h_t = {\rm Transformer_{TIR}}(\bm{F}^{(I)}, \bm{F}^{(R)}_{<t}); \\
&p_\theta(r_t|R_{<t}, I)={\rm softmax}({\rm Head}(h_t)).
\end{align}
Specifically, ${\rm LSU}(\cdot)$ unifies $I$ and $R$ into discrete tokens and produces their embeddings $\bm{E}^{(I)}$ and $\bm{E}^{(R)}$ respectively. ${\rm CRA}(\cdot)$ takes embeddings as the input and outputs discriminative features $\bm{F}^{(I)}$ and $\bm{F}^{(R)}$, which will be used to contrast and align with each other to enhance global cross-modal alignment (as introduced next). Next, ${\rm Transformer_{TIR}}(\cdot)$, a standard Transformer~\cite{vaswani2017attention} integrated with our proposed TIR to improve fine-grained cross-modal alignment, calculates the hidden state at the $t$-th time step ($h_t$) based on the visual features $\bm{F}^{(I)}$ and the preceding textual features $\bm{F}^{(R)}_{0:t-1}$. Finally, the hidden state $h_t$ is mapped into the distribution over the vocabulary by ${\rm Head(\cdot)}$ (\ie, a fully-connected layer and a ${\rm softmax}$ function). The basic training objective of our UAR for language modeling is defined as:
\begin{align}
\label{eq:loss_ce}
    \mathcal{L}_{CE} = -\sum_{t=1}^{T}\log p_\theta(r=r_t|R_{<t}, I)
\end{align}

In the following, we will elaborate on our proposed LSU, CRA, and TIR, followed by the introduction of a two-stage training procedure.

\subsection{Latent Space Unifier (LSU)}
\label{sec:lsu}
As shown in Figure~\ref{fig:Framework}~(a), LSU unifies image and text modalities into discrete tokens and obtain their embeddings. 

For extracting visual embeddings $\bm{E}^{(I)}$ of the input radiograph $I$, we follow~\cite{ramesh2021zero} and~\cite{baobeit} to exploit a discrete variational autoencoder (dVAE), which involves three components, \ie, a codebook $\mathcal{V}_I$, an encoder that maps an image into distributions over the codebook, and a decoder that reconstructs the image from its discrete tokens sampled from the distribution. Please see~\cite{ramesh2021zero} for more details of dVAE. Here, we only adopt the codebook and the encoder of dVAE and discard its decoder. Concretely,  the encoder of dVAE compresses $I \in \mathbb{R}^{H \times W \times C}$ into $\bm{D} \in \mathbb{R}^{L \times |\mathcal{V}_I|}$, where $L = HW/M^2$ denotes the number of visual tokens, $M$ is the downsampling factor, and $|\mathcal{V}_I|$ the size of the codebook. We apply the $\rm argmax$ operation on $\bm{D}$ to obtain $L$ visual tokens, and then use a trainable look-up matrix $\mathbf{W_I}\in \mathbb{R}^{|V_I|\times d}$ to attain $\bm{E}^{(I)} \in \mathbb{R}^{L \times d}$.

As the radiology report $R$ is already discrete data, we only need a vocabulary $\mathcal{V}_R$ (\eg, constructed from the training corpus) and a trainable look-up matrix $\mathbf{W_R} \in \mathbb{R}^{|\mathcal{V}_R| \times d}$ for extracting text embeddings $\bm{E}^{(R)} \in \mathbb{R}^{T\times d}$.


It is noteworthy that although the dVAE we use is pre-trained on non-medical images, it produces reasonable reconstruction results for X-ray images (see Appendix).

\subsection{Cross-modal Representation Aligner (CRA)}
\label{sec:cra}
As shown in Figure~\ref{fig:Framework}~(b), CRA is modality-agnostic. It processes visual and textual embeddings with the same set of orthonormal basis and a dual-gate mechanism to model cross-modal interactions, and enhances global semantic alignment under a triplet contrastive loss.

Specifically, we construct a set of orthonormal basis $\bm{B} \in \mathbb{R}^{2,048 \times d}$ via Gram-Schmidt Orthogonalization~\cite{schmidt1907theorie}. Similar to Layer Normalization~\cite{ba2016layer}, we adjust $\bm{B}$ as follows:
\begin{align}
   \hat{\bm{B}} = \bm{\gamma} \odot \bm{B} + \bm{\beta} ,
\end{align}
where $\odot$ is element-wise multiplication, $\bm{\gamma}$ and $\bm{\beta}$ are gain and bias parameters of the same dimension as $\bm{B}$. Next, we use $\hat{\bm{B}}$ and the attention mechanism to process embeddings from different modalities:
\begin{equation}
 \begin{aligned}
\label{eq:attn}
&\tilde{\bm{F}}^{(*)} = \text{Attention}(\bm{E}^{(*)}, \hat{\bm{B}}, \hat{\bm{B}}),\\
&\text{Attention}(\bm{Q},\bm{K},\bm{V}) = \text{softmax}(\bm{A})(\bm{V}\textbf{W}_V),\\
&\bm{A} = (\bm{Q}\textbf{W}_Q)(\bm{K}\textbf{W}_K)^\top/\sqrt{d},
\end{aligned}   
\end{equation}
where $* \in \{I, R\}$, $\bm{A}$ denotes attention activations, and all $\textbf{W}$ are trainable parameters. In practice, we extend the above attention mechanism into a multi-head version as in \cite{vaswani2017attention}. By representing features with the same basis $\hat{\bm{B}}$, it could benefit feature alignment~\cite{wu1995orthogonal, strang1999discrete}. Next, inspired by LSTM~\cite{hochreiter1997long}, we define a gate mechanism to control the information flow:
\begin{align}
    G(\bm{X}, \bm{Y}) = \sigma(\bm{X}\textbf{W}_1 + \bm{Y}\textbf{W}_2),
\end{align}
where $\sigma$ is the sigmoid function, $\textbf{W}_1$ and $\textbf{W}_2$ are trainable weights. We introduce a input gate $G_I(\cdot,\cdot)$ and a forget gate $G_F(\cdot,\cdot)$ to adaptively fuse $\bm{E}^{(*)}$ and $\tilde{\bm{F}}^{(*)}$ to produce the final features $\bm{F}^{(*)}$ as follows:

\begin{equation}
\begin{aligned}
    \bm{F}^{(*)} = \;
    &G_I(\bm{E}^{(*)}, \tilde{\bm{F}}^{(*)})
    \odot 
    \tanh{(\bm{E}^{(*)} + \tilde{\bm{F}}^{(*)})} 
    \\+ \;
    &G_F(\bm{E}^{(*)}, \tilde{\bm{F}}^{(*)})
    \odot 
    \bm{E}^{(*)} 
    + 
    \tilde{\bm{F}}^{(*)}.
\end{aligned}    
\end{equation}

\smallskip\noindent\textbf{Triplet Contrastive Loss} \
In order to achieve global cross-modal alignment with explicit supervision signals, we introduce the following training objective: 
\begin{equation}
\label{eq:triplet_loss}
\small
\begin{aligned}
    \mathcal{L}_{Global} = \;
    &{\rm ReLU} \left(\alpha - \langle \bm{F}^{(I)}, \bm{F}^{(R)}\rangle +  \langle \bm{F}^{(I)},  \bm{F}^{(R)}_-\rangle\right) \\ + \; 
    &{\rm ReLU} \left(\alpha - \langle \bm{F}^{(I)}, \bm{F}^{(R)}\rangle +  \langle \bm{F}^{(I)}_-,  \bm{F}^{(R)}\rangle\right)
\end{aligned}
\end{equation}
where $\langle \cdot, \cdot \rangle$ denotes cosine similarity, $\alpha$ is the margin value, and $\bm{F}^{(R)}_-$ and $\bm{F}^{(I)}_-$ indicate hard negative samples that have a high cosine similarity with the anchors $\bm{F}^{(I)}$ and $\bm{F}^{(R)}$. 

\subsection{Text-to-Image Refiner (TIR)}
\label{sec:ft2tr}
TIR aims to enhance the correspondence between the radiology report $R$ and its associated image $I$ at the token level. This is achieved by affects text-to-image attention patterns with an additional learnable mask, as shown in Figure~\ref{fig:Framework}~(c) and (e). Specifically, we modify the definition of $\bm{A}$ (Eq.~\ref{eq:attn}) in the vanilla text-to-image attention of the Transformer decoder as follows:
\begin{align}
\label{eq:t2tattn}
\bm{A} &= \left((\bm{Q}\textbf{W}_Q)(\bm{K}\textbf{W}_K)^\top + k \cdot \sigma(\bm{M})\right)/\sqrt{d}
\end{align}
where $k$ is a large scaling constant (\eg, 1,000) and $\sigma(\bm{M}) \in \mathbb{R}^{T\times L}$ denotes the learnable mask that impacts the activations between $T$ textual tokens and $L$ visual tokens. To focus each text token on proper visual information, we constrain the mask with the following objective:
\begin{align}
\label{eq:loss_mask}
    \mathcal{L}_{Mask} = \sum_{i=1}^T\sum_{j=1}^L (1 - \sigma(\bm{M}_{ij})).
\end{align}

\begin{table*}[ht]
\footnotesize
\begin{center}
\begin{tabular}{@{}l l c c c c c c c@{}}
\toprule
\textsc{\textbf{Dataset}}& \textsc{\textbf{Method}} & \textbf{BLEU-1} & \textbf{BLEU-2} & \textbf{BLEU-3} & \textbf{BLEU-4} & \textbf{METEOR}  & \textbf{ROUGE-L} & \textbf{\textsc{CIDEr}} \\
\midrule 
\multirow{11}{*}[-3pt]{\textsc{IU-Xray}~\cite{demner2016preparing}} & \textsc{CoAtt~\cite{jing2018automatic}}  & $0.455$ & $0.288$ & $0.205$ & $0.154$ & - & $0.369$ & $0.277$ \\
& HRGR~\cite{li2018hybrid}  & $0.438$ & $0.298$ & $0.208$ & $0.151$ & - & $0.322$ & $0.343$ \\
& CMAS-RL~\cite{jing2019show} & $0.464$ & $0.301$ & $0.210$ & $0.154$ & - & $0.362$ & $0.275$ \\
& \textsc{SentSAT+KG}~\cite{zhang2020radiology} & $0.441$ & $0.291$ & $0.203$ & $0.417$ & - & $0.304$ & $0.304$ \\
& \textsc{R2Gen~\cite{chen2020generating}} & $0.470$ & $0.304$ & $0.219$ & $0.165$ & - & $0.371$ & - \\
& CMCL~\cite{liu2021competence} & $0.473$ & $0.305$ & $0.217$ & $0.162$ & $0.186$ & $0.378$ & - \\
& PPKED~\cite{liu2021exploring} & $0.483$ & $0.315$ & $0.224$ & $0.168$ & $0.190$ & $0.376$ & $\colorbox{green!15}{0.351}$ \\
& \textsc{R2GenCMN*}~\cite{chen2021cross} & $0.470$ & $0.304$ & $0.222$ & $0.170$ & $0.191$ & $0.358$ & $0.344$ \\
& JPG~\cite{you2022jpg} & $0.479$ & $0.319$ & $0.222$ & $0.174$ & $0.193$ & $0.377$ & - \\
& CMM+RL~\cite{qin2022reinforced} & $\colorbox{green!15}{0.494}$ & $\colorbox{green!15}{0.321}$ & $\colorbox{green!15}{0.235}$ & $\colorbox{green!15}{0.181}$ & $\colorbox{green!15}{0.201}$ & $\colorbox{green!15}{0.384}$ & - \\
\cmidrule(l){2-9}
& UAR (\textbf{Ours}) & $\colorbox{blue!15}{0.530}$ & $\colorbox{blue!15}{0.365}$ & $\colorbox{blue!15}{0.263}$ & $\colorbox{blue!15}{0.200}$& $\colorbox{blue!15}{0.218}$ & $\colorbox{blue!15}{0.405}$& $\colorbox{blue!15}{0.501}$\\
\midrule
\multirow{10}{*}[-3pt]{MIMIC-CXR~\cite{johnson2019mimic}} & ST~\cite{vinyals2015show}  & $0.299$ & $0.184$ & $0.121$ & $0.084$ & $0.124$ & $0.263$ & - \\
& \textsc{Att2in}~\cite{rennie2017self} & $0.325$ & $0.203$ & $0.136$ & $0.096$ & $0.134$ & $0.276$ & - \\
& \textsc{AdaAtt}~\cite{lu2017knowing} & $0.299$ & $0.185$ & $0.124$ & $0.088$ & $0.118$ & $0.266$ & - \\
& \textsc{TopDown}~\cite{anderson2018bottom} & $0.317$ & $0.195$ & $0.130$ & $0.092$ & $0.128$ & $0.267$ & - \\
& \textsc{R2Gen}~\cite{chen2020generating} & $0.353$ & $0.218$ & $0.145$ & $0.103$ & $0.142$ & $0.270$ & - \\
& CMCL~\cite{liu2021competence} & $0.344$ & $0.217$ & $0.140$ & $0.097$ & $0.133$ & $0.281$ & - \\
& PPKED~\cite{liu2021exploring} & $0.360$ & $0.224$ & $0.149$ & $0.106$ & $0.149$ & $0.284$ & $\colorbox{green!15}{0.237}$ \\
& \textsc{R2GenCMN*}~\cite{chen2021cross} & $0.348$ & $0.206$ & $0.135$ & $0.094$ & $0.136$ & $0.266$ & $0.158$ \\
& CMM+RL~\cite{qin2022reinforced} & $\colorbox{blue!15}{0.381}$ & $\colorbox{blue!15}{0.232}$ & $\colorbox{green!15}{0.155}$ & $\colorbox{blue!15}{0.109}$ & $\colorbox{green!15}{0.151}$ & $\colorbox{green!15}{0.287}$ & - \\
\cmidrule(l){2-9}
 & UAR (\textbf{Ours})  & $\colorbox{green!15}{0.363}$ & $\colorbox{green!15}{0.229}$ & $\colorbox{blue!15}{0.158}$ & $\colorbox{green!15}{0.107}$ & $\colorbox{blue!15}{0.157}$ & $\colorbox{blue!15}{0.289}$ & $\colorbox{blue!15}{0.246}$\\
\bottomrule
\end{tabular}
\end{center}
\caption{Comparison with state-of-the-art methods on IU-Xray and MIMIC-CXR datasets. \colorbox{blue!15}{Optimal} and \colorbox{green!15}{suboptimal} performance is highlighted. *: Our re-implementations with the official code. \textbf{Our UAR achieves competitive if not the best performance on all metrics.}}
\label{tab:main_reslut}
\end{table*}

\subsection{Two-Stage Training}
\label{sec:moo}
So far, we have introduced three training objectives of our UAR approach: $\mathcal{L}_{CE}$ (Eq.~\ref{eq:loss_ce}) for language modeling, $\mathcal{L}_{Global}$ (Eq.~\ref{eq:triplet_loss}) for enhancing global cross-modal alignment, and $\mathcal{L}_{Mask}$ (Eq.~\ref{eq:loss_mask}) that re-calibrates attention activations to boost local cross-modal alignment. The overall loss function of our approach is formulated as:
\begin{align}
    \label{eq:loss_overall}
    \mathcal{L} = \lambda_1\mathcal{L}_{CE} + \lambda_2\mathcal{L}_{global} + \lambda_3\mathcal{L}_{Mask}.
\end{align}
To learn cross-modal alignment incrementally and stably, we split the training process into two stages. In the first stage, $\{\lambda_1, \lambda_2, \lambda_3\} = \{1, 1, 0\}$. The model does not include a learnable mask in the text-to-image attention and is forced to learn coarse-grained semantic alignment besides language modeling. In the second stage, $\{\lambda_1, \lambda_2, \lambda_3\} = \{1, 1, 1\}$, \ie, the model additionally emphasizes the learning of fine-grained semantic alignment. 

\normalem \section{Experiments}

\subsection{Configurations}
\smallskip\noindent\textbf{Datasets} \
Our evaluation is performed on two publicly available radiology report generation benchmark datasets, namely IU-Xray~\cite{demner2016preparing} and MIMIC-CXR~\cite{johnson2019mimic}.

$\bullet$  \textbf{IU-Xray}\footnote{\url{https://openi.nlm.nih.gov/faq/}}, established by Indiana University, is a commonly-used dataset that comprises $7,470$ X-ray images along with $3,955$ corresponding reports. 
We use the same $70\%$-$10\%$-$20\%$ training-validation-testing splits as previous works~\cite{chen2021cross, liu2021exploring, you2022jpg}. 
We keep words that appear more than 3 times, resulting a vocabulary of size around 1K.

$\bullet$  \textbf{MIMIC-CXR}\footnote{\url{https://physionet.org/content/mimic-cxr/}}, provided by Beth Israel Deaconess Medical Center, is a recently released large chest X-ray dataset that contains $473,057$ radiographs and $206,563$ corresponding reports. We adopt the official splits and retain words that appear more than 10 times, resulting in a vocabulary of roughly 4K words.

\smallskip\noindent\textbf{Metrics} \
We report four widespread automatic metrics: BLEU~\cite{papineni2002bleu}, METEOR~\cite{banerjee2005meteor}, ROUGE-L~\cite{lin2004rouge} and CIDEr~\cite{vedantam2015cider}, and compute them with Microsoft COCO Evaluation Server~\cite{chen2015microsoft}. Higher is better for all metrics.

\smallskip\noindent\textbf{Settings} \
Our baseline model comprises a pre-trained ResNet101~\cite{he2016deep} and a randomly initialized Transformer-Base~\cite{vaswani2017attention} with 3 layers. For our UAR model, we replace ResNet101 with a pre-trained dVAE, whose details are left to Appendix. Following previous works~\cite{li2018hybrid, liu2021exploring, you2022jpg}, we use frontal and lateral X-ray images as input on IU-Xray, and only frontal X-ray images on MIMIC-CXR. To save computation cost, we resize images to $128 \times 128$ and randomly crop a region of size $112 \times 112$ at the training phase; we directly resize the image to $112 \times 112$ at the inference phase. We use the AdamW optimizer~\cite{loshchilov2017decoupled} to train the model. We employ distinct training stage arrangements and learning rate schedules, along with varying batch sizes, for different datasets. More implementation details are given in Appendix. 

\begin{table*}[ht]

\footnotesize
\setlength{\tabcolsep}{2.5pt}

\begin{center}
\begin{tabular}{@{}c c c c c c c   c c c c c c c@{}}
\toprule
\multirow{2}{*}[-3pt]{\textbf{\textsc{Section}}} & \multirow{2}{*}[-3pt]{\textbf{\textsc{Model}}}  &\multirow{2}{*}[-3pt]{\textbf{\textsc{LSU}}} & \multicolumn{2}{c}{\textbf{\textsc{CRA}}}  & \multirow{2}{*}[-3pt]{\textbf{\textsc{TIR}}} & 
\multirow{2}{*}[-3pt]{\begin{tabular}[c]{@{}c@{}}  \textbf{\textsc{Two-Stage}}\\\textbf{\textsc{Training}}\end{tabular}} &
\multicolumn{7}{c}{\textbf{\textsc{Dataset: IU-Xray }}~\cite{demner2016preparing}}     \\

\cmidrule(lr){4-5} \cmidrule(lr){8-14} & & & Integration & $\mathcal{L}_{global}$ && & \textsc{BLEU-1} & \textsc{BLEU-2} & \textsc{BLEU-3} & \textsc{BLEU-4} & METEOR & ROUGE-L & \textsc{CIDEr}  \\ \midrule [\heavyrulewidth]

\multirow{2}{*}{\textit{\ref{sec:lsu}}}  & \textsc{Base}  & & & & 
& & $0.424$ & $0.268$ & $0.189$ & $0.144$ & $0.171$ & $0.339$ & $0.271$
\\
& (a)  & $\surd$ & & & &  & $0.475$ & $0.319$ & $0.233$ & $0.177$ & $0.218$ & $0.399$ & $0.295$ \\ \midrule

\multirow{3}{*}{\textit{\ref{sec:cra}}} & (b)  & & $\surd$ & &  
& & $0.446$ & $0.283$ & $0.196$ & $0.141$ & $0.174$ & $0.360$ & $0.330$ \\

& (c)  & & & $\surd$ & & 
 & $0.426$ & $0.274$ & $0.192$ & $0.143$ & $0.184$ & $0.356$ & $0.409$ \\

& (d)  & & $\surd$ & $\surd$ & & 
 & $0.471$ & $0.302$ & $0.213$ & $0.158$ & $0.187$ & $0.390$ & $0.393$ \\ \midrule

\multirow{2}{*}{\textit{\ref{sec:ft2tr}}} & (e)  & $\surd$ & $\surd$ & $\surd$ & & 
 & $\colorbox{blue!15}{0.538}$ & $\colorbox{green!15}{0.365}$ & $\colorbox{green!15}{0.262}$ & $\colorbox{green!15}{0.193}$ & $\colorbox{green!15}{0.218}$ & $0.365$ & $0.410$  \\

& (f) & $\surd$ & $\surd$ & $\surd$ & $\surd$
 &  & $0.505$ &  $0.332$ &  $0.244$ &  $0.182$ &  $0.206$&  $\colorbox{green!15}{0.401}$  &  $\colorbox{green!15}{0.460}$ \\
\midrule
\textit{\ref{sec:moo}}& UAR (\textbf{Ours})  & $\surd$ & $\surd$ & $\surd$ & $\surd$
 & $\surd$ &  $\colorbox{green!15}{0.530}$ &  $\colorbox{blue!15}{0.365}$ &  $\colorbox{blue!15}{0.263}$ &  $\colorbox{blue!15}{0.200}$ &  $\colorbox{blue!15}{0.218}$ &  $\colorbox{blue!15}{0.405}$ &  $\colorbox{blue!15}{0.501}$
\\
\bottomrule
\end{tabular}
\end{center}
\caption{Ablation studies on the proposed Latent Space Unifier (LSU), Cross-modal Representation Aligner (CRA), Text-to-Image Refiner (TIR), and the two-stage training procedure. In CRA, ``Integration'' means using the orthogonal subspace and the dual-gate mechanism simultaneously. As we can observe, \textbf{our UAR model significantly outperforms the base model on all metrics}.}
\label{tab:ablation}
\end{table*}

\begin{table}[ht]
    \renewcommand\arraystretch{1.4}
    \setlength{\tabcolsep}{2.5pt}
    \centering
    \footnotesize  
    \begin{center}
    \begin{tabular}{@{}l c c c c c c c@{}}
        \toprule
        & BL-1 & BL-2 & BL-3 & BL-4 & MTOR & RG-L & \textsc{CIDEr} \\ \midrule [\heavyrulewidth]

        \textbf{Ours}
        & $\colorbox{blue!15}{0.530}$ & $\colorbox{blue!15}{0.365}$ & $\colorbox{blue!15}{0.263}$ & $\colorbox{blue!15}{0.200}$ & $\colorbox{blue!15}{0.218}$ & $\colorbox{blue!15}{0.405}$ & $\colorbox{blue!15}{0.501}$ \\
        
        Uniform & 0.441 & 0.284 & 0.207 & $\colorbox{green!15}{0.158}$ & 0.179 & $\colorbox{green!15}{0.378}$ & $\colorbox{green!15}{0.428} $\\
        Normal & $\colorbox{green!15}{0.467}$ & $\colorbox{green!15}{0.303}$ & $\colorbox{green!15}{0.210}$ & 0.155 & $\colorbox{green!15}{0.197}$ & 0.371 & ${0.427}$ \\ 
        \bottomrule
    \end{tabular}
    \end{center}
    \caption{Effect of using subspaces subject to \textbf{orthogonal (default)}, uniform, and normal distributions in the proposed CRA.
    \label{tab:basis}}
\end{table}

\subsection{Comparison with State-of-the-Art Methods}

We compare our UAR model with state-of-the-art (SOTA) methods of different backbones, including CNN-RNN~\cite{vinyals2015show, lu2017knowing, anderson2018bottom, jing2018automatic} and CNN-Transformer~\cite{chen2020generating, chen2021cross, liu2021exploring, you2022jpg}. 
Besides, we also consider SOTA methods using different technologies, including reinforcement learning~\cite{rennie2017self, li2018hybrid, jing2019show, qin2022reinforced}, knowledge graph~\cite{zhang2020radiology}, and curriculum learning~\cite{liu2021competence}. As shown in Table~\ref{tab:main_reslut}, our approach achieves SOTA performance on the IU-Xray dataset, with up to 1.9\% and 15\% absolute gains in terms of BLEU-4 and CIDEr metrics. 
As for the MIMIC-CXR dataset, our UAR obtains competitive results compared to the strongest competitor, \ie, CMM+RL~\cite{qin2022reinforced}, which utilizes reinforcement learning to directly optimize metric scores. 
Notably, as our approach mainly involves network designing, incorporating existing technologies like reinforcement learning and curriculum learning may bring further improvements. 
Besides, the superiority of our approach against CoATT~\cite{jing2018automatic}, R2GenCMN~\cite{chen2021cross}, and JPG~\cite{you2022jpg} show the necessity of capturing multi-level cross-modal alignments.

\normalem \section{Ablation Study and Analysis}
In this section, we delve deeper into each module of our UAR with both quantitative and qualitative results.

\subsection{Effect of Latent Space Unifier}
\begin{figure}[ht]
\begin{center}
\includegraphics[width=1\linewidth]{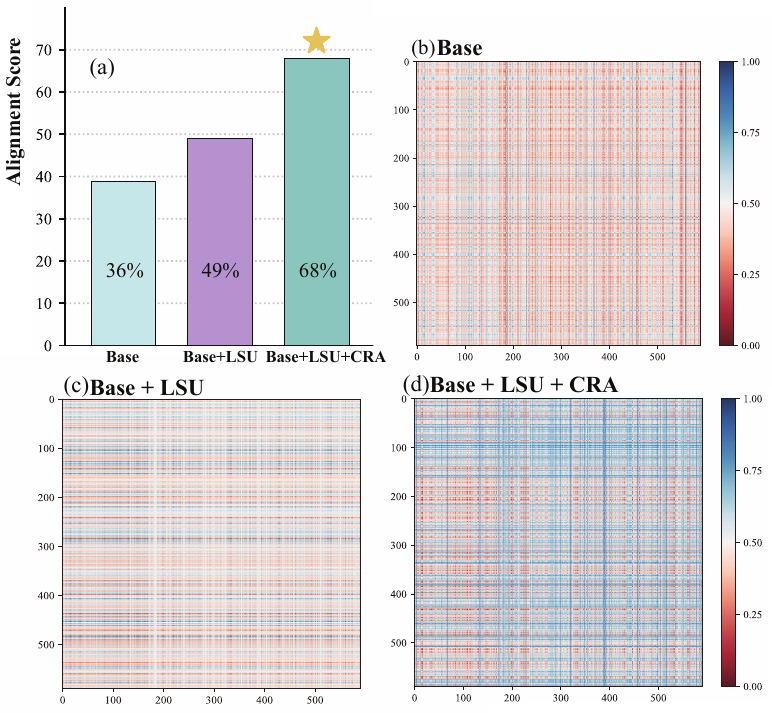}
\end{center}
\caption{(a): Alignment scores of different models. (b-d): Heatmaps of pairwise cosine similarity among all data samples. }
\label{fig:aligned_socre}
\end{figure}

\smallskip\noindent\textbf{Quantitative Analysis} \
As shown in Table~\ref{tab:ablation} show, model (a) surpasses the base model on all metrics, \eg, $2.3$\% improvements on BLEU-4. 
Meanwhile, model (a) is also a strong competitor even compared with the SOTA methods listed in \cref{tab:main_reslut} on IU-Xray. 
The performance boost is possibly because LSU reconciles the distributions of visual and textual data and thus prompts the implicit learning of cross-modal alignments. 
To verify our hypothesis, we calculate the \textit{alignment score}\footnote{Alignment score is defined as the portion of radiograph-report pairs whose cosine similarity of features is larger than 0.5 after min-max normalization.} of different models. As shown in Figure~\ref{fig:aligned_socre}~(a), we can observe that ``Base + LSU'' achieves higher alignment score than the base model, \ie, $49$\% vs. $36$\%. which proves the potential of LSU in implicitly aligning features of different modalities.

\begin{figure*}[ht]
\begin{center}
\includegraphics[width=1\linewidth]{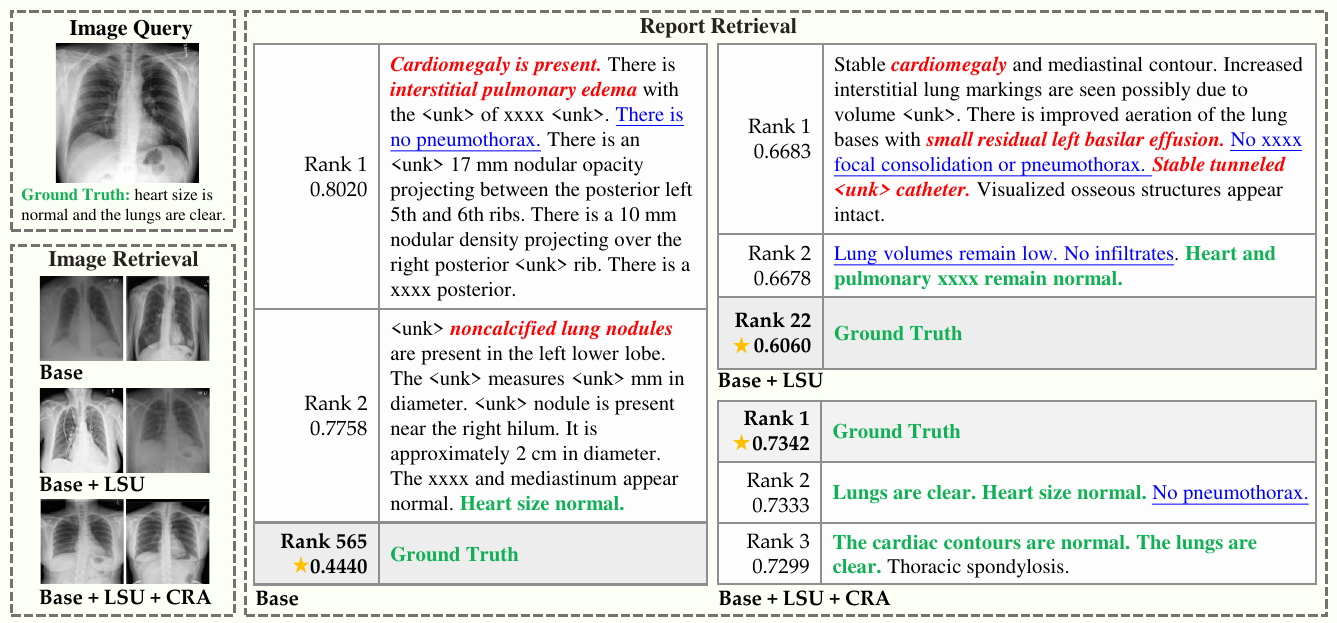}
\end{center}
\caption{Image retrieval and report retrieval results of different models. ``Base'', ``Base + LSU'', and ``Base + LSU + CRA'' correspond to the base model, model (a) and model (e) in \cref{tab:ablation}, respectively. In report retrieval, we highlight \textcolor[RGB]{16,177,84}{\textbf{accurate}}, \textcolor[RGB]{0,0,255}{\uline{reasonable}}, and \textcolor{red}{{\textit{\textbf{wrong}}}} phrases. We also foreground rankings and cosine similarities. As we can observe, the model is more likely to retrieve the ground-truth report by gradually integrating the base model with our proposed LSU and CRA, \ie, \textbf{our approach effectively enhances global semantic alignments}.
\label{fig:ua_abl}}
\end{figure*}

\smallskip\noindent\textbf{Qualitative Analysis} \
We visualize the heatmaps of pairwise cosine similarity among all data samples in Figure~\ref{fig:aligned_socre} (b-d). 
As we can see in (b), samples similar to the query sample are relatively rare in the base model. 
Considering that radiology reports contain inherent patterns, high correlations shall be observed frequently. 
In (c), integrating the base model with LSU improves this situation to some extent via unifying multimodal characteristics. 
Furthermore, Figure~\ref{fig:ua_abl} shows retrieval results of different models given the same image as the query. 
We can see that compared with the base model, ``Base + LSU'' gives higher cosine similarity between the query image and the ground-truth report, \ie, $0.4440\rightarrow 0.6060$. These results further confirm that LSU can benefit global semantic alignments.

\begin{table}[t]
    \centering
    \footnotesize  
    \renewcommand\arraystretch{1.4}
    \setlength{\tabcolsep}{2.5pt}
    \begin{center}
    \begin{tabular}{@{}l c c c c c c c@{}}
        \toprule
        & BL-1 & BL-2 & BL-3 & BL-4 & MTOR & RG-L & \textsc{CIDEr} \\ \midrule [\heavyrulewidth]
        \textbf{Ours}  & $\colorbox{blue!15}{0.530}$ & $\colorbox{blue!15}{0.365}$ & $\colorbox{blue!15}{0.263}$ & $\colorbox{blue!15}{0.200}$ & $\colorbox{blue!15}{0.218}$ & $\colorbox{blue!15}{0.405}$ & $\colorbox{blue!15}{0.501}$  \\
        
        No Gate & 0.439 & 0.296 & 0.208 & 0.155 & 0.193 & 0.373 & $\colorbox{green!15}{0.405}$\\
        Addition & $\colorbox{green!15}{0.487}$ & $\colorbox{green!15}{0.318}$ & $\colorbox{green!15}{0.225}$ & $\colorbox{green!15}{0.167}$ & $\colorbox{green!15}{0.214}$ & $\colorbox{green!15}{0.386}$ & 0.376 \\ 
        
        \bottomrule
    \end{tabular}
    \end{center}
    \caption{Effect of different fusing mechanisms in the proposed CRA, including \textbf{dual-gate (default)}, ``no gate'': $\bm{F}^{(*)} = \tilde{\bm{F}}^{(*)}$, and ``addition'': $\bm{F}^{(*)} = \bm{E}^{(*)} + \tilde{\bm{F}}^{(*)}$.
    \label{tab:integration}}
\end{table}

\begin{figure}[ht]
\begin{center}
\includegraphics[width=0.7\linewidth]{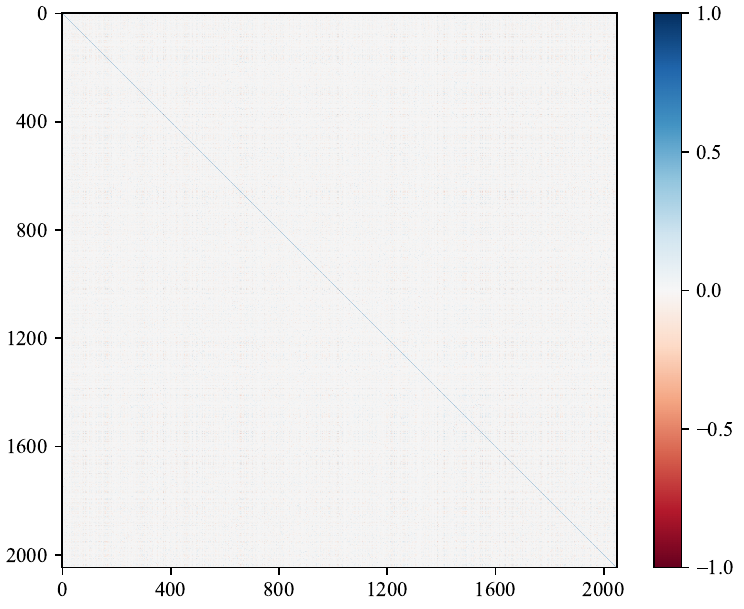}
\end{center}
\caption{
Visualization of Gram matrix for the memory matrix trained in R2GenCMN~\cite{chen2021cross}.
\label{fig:r2gencmn_subspace}}
\end{figure}

\subsection{Effect of Cross-modal Representation Aligner}
\smallskip\noindent\textbf{Quantitative Analysis} \
By comparing models (b,c,d) with the base model in Table~\ref{tab:ablation}, we can observe that both the feature integration and the triplet contrastive loss of CRA can bring performance boosts. 
Specifically, the feature integration encourages the model to generate fluent and faithful reports and thus model (b) achieves higher BLEU and ROUGE-L scores than the base model. 
Meanwhile, the triplet contrastive loss promotes the generation of diverse sentences that are semantically consistent with ground truth report, resulting in higher CIDEr score of model (c) than the base model. 
By combining complementary merits of models (b) and (c), model (d) performs better. 
Moreover, compared with model (d), incorporating LSU and CRA (\ie, model (e)) brings furhter improvements, which can be attributed to the better cross-modal alignment ability shown in Figure~\ref{fig:aligned_socre} (a), where model (e) (\ie, ``Base + LSU + CRA'') gets the highest alignment score (68\%). 

As the feature integration of CRA is composed of two parts: the orthogonal subspace~\footnote{Since we use the scaled mechanism like Layer Normalization~\cite{ba2016layer} to adjust the basis, the orthonormal subspace evolved into orthogonal.} and the two-gate mechanism, we investigate different variants that can be substituted with our proposals in \cref{tab:basis} and \cref{tab:integration}. We can see that our proposed orthogonal subspace outperforms the uniform and normal ones, and the dual-gate mechanism is superior in feature fusion than all other variants.

\begin{figure*}[!ht]
\begin{center}
\includegraphics[width=0.95\linewidth]{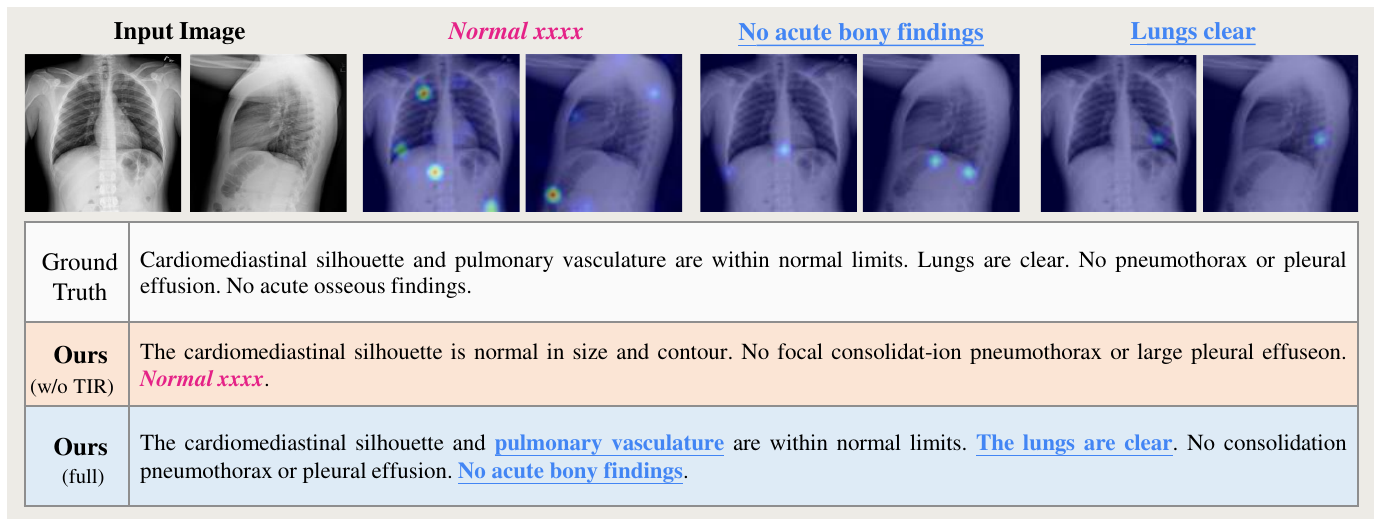}
\end{center}
\caption{An example of radiology reports generated by our model without or with the proposed Text-to-Image Refiner (TIR). We highlight \textcolor[RGB]{229, 34, 135}{\textit{\textbf{incomplete}}} and \textcolor[RGB]{66, 133, 244}{\textbf{\uline{detailed}}} descriptions in the report. We also show the text-to-image attention heatmaps of specific descriptions at the top. \textbf{Our full model learns accurate correspondences between image regions and keywords.}}
\label{fig:TIR}
\end{figure*}

\smallskip\noindent\textbf{Qualitative Analysis} \
Here, we first probe into the SOTA R2GenCMN model~\cite{chen2021cross} that enhance global semantic alignments with a memory matrix. We visualize the Gram matrix of the pre-trained memory matrix in Figure~\ref{fig:r2gencmn_subspace}. Surprisingly, almost all values in Gram matrix equal to 0, demonstrating that it is approximately an orthogonal subspace. This visualization result justifies our subspace design in CRA.

Likewise, we show the similarity heatmap of ``Base + LSU + CRA" in Figure~\ref{fig:aligned_socre} (d), whose pattern is quite different to that of (b) and (c). In Figure~\ref{fig:ua_abl}, we can observe that ``Base + LSU + CRA" performs the best in retrieval. 
On the one hand, retrieved reports have high similarity to the query image, and they are semantically consistent with the ground-truth report, \eg, \textit{lung are clear} and \textit{the cardiac contours are normal}. 
On the other hand, the ground-truth report can be retrieved precisely, \ie, ranking first with a similarity of 0.7342.

\par In brief, the above quantitative and qualitative analyses verify that our proposed LSU and CRA can effectively enhances global semantic alignments.

\subsection{Effect of Text-to-Image Refiner}
\smallskip\noindent\textbf{Quantitative Analysis} \
By comparing model (e), (f), and our UAR model in \cref{tab:ablation}, we have the following observations. 1) TIR can effectively improves the CIDEr metric, \eg, 5\% absolute improvement for model (f) vs. model (e), showing that TIR is beneficial for generating accurate keywords in the radiology report. Moreover, excluding two-stage training leads to obvious performance degradation, \ie, our UAR model vs. model (f). This indicates the instability of multi-objective optimization and proves the effectiveness of our two-stage training procedure.  

\smallskip\noindent\textbf{Qualitative Analysis} \
In Figure~\ref{fig:TIR}, we illustrate the effect of TIR on radiology report generation.
As we can see, the model without TIR outputs an useless description ``Normal xxxx'', where ``xxxx`` is anonymized information used to protect patient privacy and results in an incomplete sentence.
By contrast, our TIR remedies this content omission and interprets the visual content in detail. 
Additionally, text-to-image attention heatmaps demonstrate that our TIR is capable of focusing word prediction on proper visual information.

In all, TIR can capture accurate fine-grained correspondences between image regions and keywords. Our full model can generate more informative and meaningful radiology reports by enhancing multi-level semantic alignments.

\normalem \section{Conclusion}
In this paper, we make the first attempt to align visual and textual semantics at different levels with explicit constraints in automatic radiology report generation. To this end, we propose the UAR approach to unify multimodal data into discrete tokens, align visual and textual representations globally, and refine fine-grained text-to-image correspondences. Extensive experiments and analyses on two benchmark datasets, \ie, IU-Xray and MIMIC-CXR, demonstrate that our approach can generate more informative and meaningful radiology reports by boosting multi-level cross-modal alignments and thus achieves state-of-the-art performance. 

{\small
\normalem
\bibliographystyle{ieee_fullname}
\bibliography{egbib}
}

\end{document}